\documentclass{article}

\usepackage{CJKutf8} 
\usepackage[utf8]{inputenc}

\usepackage[preprint,nonatbib]{neurips_2022}

\usepackage[utf8]{inputenc} 
\usepackage[T1]{fontenc}    
\usepackage{hyperref}       
\usepackage{url}            
\usepackage{booktabs}       
\usepackage{amsfonts}       
\usepackage{nicefrac}       
\usepackage{microtype}      
\usepackage{xcolor}         

\usepackage{wrapfig}
\usepackage{times}
\usepackage{latexsym}

\definecolor{darkblue}{rgb}{0.0,0.0,0.5}

\usepackage{amsmath}
\usepackage{array}
\newcolumntype{L}{>{$}l<{$}}
\newcolumntype{C}{>{$}c<{$}}
\newcolumntype{R}{>{$}r<{$}}

\usepackage{multirow}
\usepackage{tabularx, caption, boldline}
\usepackage{graphicx}
\usepackage{cellspace}
\usepackage{algpseudocode}  
\usepackage{multicol}  
\usepackage{flexisym}

\usepackage{microtype}

\usepackage{dsfont}

\makeatletter
\def\hlinewd#1{%
\noalign{\ifnum0=`}\fi\hrule \@height #1 %
\futurelet\reserved@a\@xhline}
\makeatother

\usepackage{times}
\usepackage{latexsym}

\usepackage{algpseudocode}  
\usepackage{amsmath}
\usepackage{multicol}  
\usepackage{multirow} 
\usepackage{graphicx}  
\usepackage{array}
\usepackage{arydshln}
\usepackage{booktabs}
\usepackage{textcomp}

\usepackage{amssymb}
\usepackage{pifont}

\usepackage{cleveref}
\crefformat{section}{\S#2#1#3} 
\crefformat{subsection}{\S#2#1#3}
\crefformat{subsubsection}{\S#2#1#3}

\usepackage{microtype}

\usepackage{adjustbox}
\usepackage{multicol}  
\usepackage{multirow} 
\usepackage{amsmath}
\usepackage{amsfonts}
\usepackage{makecell}
\usepackage{algpseudocode} 
\usepackage{verbatim}
\usepackage{mathtools}

\usepackage{booktabs}

\usepackage{makecell}
\usepackage{cleveref}
\usepackage[normalem]{ulem}

\usepackage{CJKutf8}

\usepackage{times}
\usepackage{latexsym}

\usepackage{footnote}
\makesavenoteenv{tabular}
\makesavenoteenv{table}

\usepackage[hang,flushmargin]{footmisc}

\usepackage[linesnumbered,ruled]{algorithm2e}
\SetAlFnt{\small}
\SetAlCapNameFnt{\normalsize}
\makeatletter
\newcommand{\nosemic}{\renewcommand{\@endalgocfline}{\relax}}
\newcommand{\dosemic}{\renewcommand{\@endalgocfline}{\algocf@endline}}
\let\oldnl\nl
\newcommand{\nonl}{\renewcommand{\nl}{\let\nl\oldnl}}
\makeatother

\usepackage{enumitem}
\usepackage{tablefootnote}

\usepackage{tabularx}
\makeatletter
\def\hlinewd#1{%
\noalign{\ifnum0=`}\fi\hrule \@height #1 %
\futurelet\reserved@a\@xhline}
\makeatother

\usepackage{minitoc}

\usepackage{cleveref}
\crefformat{section}{\S#2#1#3}
\crefformat{subsection}{\S#2#1#3}
\crefformat{subsubsection}{\S#2#1#3}
\crefrangeformat{section}{\S\S#3#1#4 to~#5#2#6}
\crefmultiformat{section}{\S\S#2#1#3}{ and~#2#1#3}{, #2#1#3}{ and~#2#1#3}
\Crefformat{figure}{#2Figure~#1#3}
\Crefmultiformat{figure}{Figures.~#2#1#3}{ and~#2#1#3}{, #2#1#3}{ and~#2#1#3}
\Crefformat{table}{#2Table~#1#3}
\Crefmultiformat{table}{Tables~#2#1#3}{ and~#2#1#3}{, #2#1#3}{ and~#2#1#3}
\Crefformat{appendix}{Appx.~\S#2#1#3}
\crefformat{algorithm}{Alg.~#2#1#3}

\definecolor{NavyBlue}{rgb}{0.1, 0.4, 0.8}
\hypersetup{colorlinks = true, linkcolor = NavyBlue,
            urlcolor  = gray,
            citecolor = NavyBlue,
            anchorcolor = NavyBlue}

\title{Unified Mind Model: \\ Reimagining Autonomous Agents in the LLM Era}

\author{
 \textbf{Pengbo Hu}$^1$\thanks{Equal Contribution}\quad
 \textbf{Xiang Ying}$^1$\footnotemark[1] \thanks{Project Lead} 
 \\
 \\
 $^1$Mindverse.ai \ \ \ \ \ 
}

\begin{document}

\maketitle

\doparttoc 
\faketableofcontents

\begin{abstract}
    
Large language models (LLMs) have recently demonstrated remarkable capabilities across domains, tasks, and languages (e.g., ChatGPT and GPT-4), reviving the research of general autonomous agents with human-like cognitive abilities. Such human-level agents require semantic comprehension and 
instruction-following capabilities, which exactly fall into the strengths of LLMs. Although there have been several initial attempts to build human-level agents based on LLMs, the theoretical foundation remains a challenging open problem. In this paper, we propose a novel theoretical cognitive architecture, the Unified Mind Model (UMM), which offers guidance to facilitate the rapid creation of autonomous agents with human-level cognitive abilities. Specifically, our UMM starts with the global workspace theory and further leverage LLMs to enable the agent with various cognitive abilities, such as multi-modal perception, planning, reasoning, tool use, learning, memory, reflection and motivation. Building upon UMM, we then develop an agent-building engine, MindOS, which allows users to quickly create domain-/task-specific autonomous agents without any programming effort \footnote{This article was completed in March 2023, but was not released earlier due to commercial issues. We are making it available today in hopes that it can still provide some inspiration to the LLM Agent community. (pbhu@mail.ustc.edu.cn)}.
\end{abstract}
 


\section{Introduction}
\label{sec:introduction}

The field of artificial intelligence has been revolutionized by the development of foundational models, particularly large language models. These models, such as ChatGPT \cite{openai2023chat,ouyang2022training} and GPT-4 \cite{openai2023gpt4,bubeck2023sparks}, have demonstrated human-level performance in various tasks. The success of these models is attributed to their vast knowledge of the world, robust semantic understanding, and strong ability to follow instructions. 
Recently, ChatGPT, along with other similar models, has been widely adopted in various fields, providing substantial support for domain-specific innovation. 

LLMs are a promising building block to support various human-level cognitive abilities for autonomous agents, and several attempts have been made in this regard.
(1) Multimodal perception. The augmentation of language models with additional modalities has garnered significant interest within the multimodal research community. Using tools that facilitate alignment between various modalities and textual modalities, these methods have demonstrated a remarkable capacity for multimodal information processing \cite{wu2023visual,shen2023hugginggpt}.
(2) Complex reason and planning. The faculty of reasoning and planning represents a fundamental cognitive capacity in humans, enabling the execution of sophisticated tasks. Numerous investigations have evidenced the potential of language models to effectively perform intricate reasoning \cite{wei2022chain,qiao2022reasoning,long2023large} and planning \cite{xie2023translating,song2022llm,huang2022language,wang2023describe} processes. 
(3) Tool use. Several works equip the language model with tool use ability, i.e. ChatGPT Plugin \cite{chatplugin2023} and HuggingGPT \cite{shen2023hugginggpt}, which allow language models to interact with the digital world. Such work demonstrates the language model's proficiency in handling complex tasks with tools, which traditionally fall under the domain of human expertise \cite{schick2023toolformer,qin2023tool,liang2023taskmatrix}.
(4) Reflection. An emerging research direction seeks to imbue language models with self-improvement \cite{chen2023teaching,madaan2023self} and reflection abilities \cite{park2023generative,shinn2023ref}. These enhancements significantly bolster the problem-solving performance of LLMs.
These works suggest that language models such as ChatGPT can function as a ``prefrontal cortex''. It potentially supports a wide range of human-like cognitive abilities.
Along this trend, language models have been a potential tool to provide a new approach and methodology for developing human-level autonomous agents.

At present, there is significant research on LLM-powered autonomous agents. We classify these agents into three types:

(1) Tool-level agents:These agents can access various tools such as web searches and APIs \cite{shen2023hugginggpt,chatplugin2023}. They help reduce barriers to knowledge acquisition and enhance problem-solving efficiency. Furthermore, they assist users in managing complex workflows by streamlining tools and actions, significantly reducing workload.

(2) Autonomous agents: These agents can perform tasks autonomously with minimal human supervision \cite{autogpt2023,baby2023agi,wang2023voyager}. They possess advanced cognitive capabilities, enabling them to use tools, plan, maintain long-term memory, and learn.

(3) Agent teams: These include agents with specialized domain knowledge in various fields. The goal is to create groups of agents that excel in domain-specific tasks \cite{nexus2023} or have defined social roles \cite{park2023generative}, thereby enhancing team coordination and cooperation to tackle complex problems effectively.

Despite the proliferation of autonomous agents, there remains a lack of unified guidelines for developing agents with human-like cognitive capabilities. Research on cognitive architectures, which began in the 1950s, aimed to create general-purpose artificial intelligence by emulating human cognitive processes \cite{laird2017standard}. These efforts, often inspired by neuroscience and psychology, offer valuable insights for building human-level autonomous agents. One promising architecture is the Global Workspace Theory (GWT), which supports macro-level collaborative computations across multiple brain regions \cite{baars1993cognitive,baars2005global,franklin2013lida}. This framework allows for the effective integration of LLM-powered function modules, enabling the successful execution of complex tasks. Thus, we propose that GWT serves as a robust macro architecture for organizing distinct functional modules, facilitating the construction of advanced autonomous agents.

\begin{figure}[!t] 
    \centering
    \includegraphics[width=0.8\columnwidth]{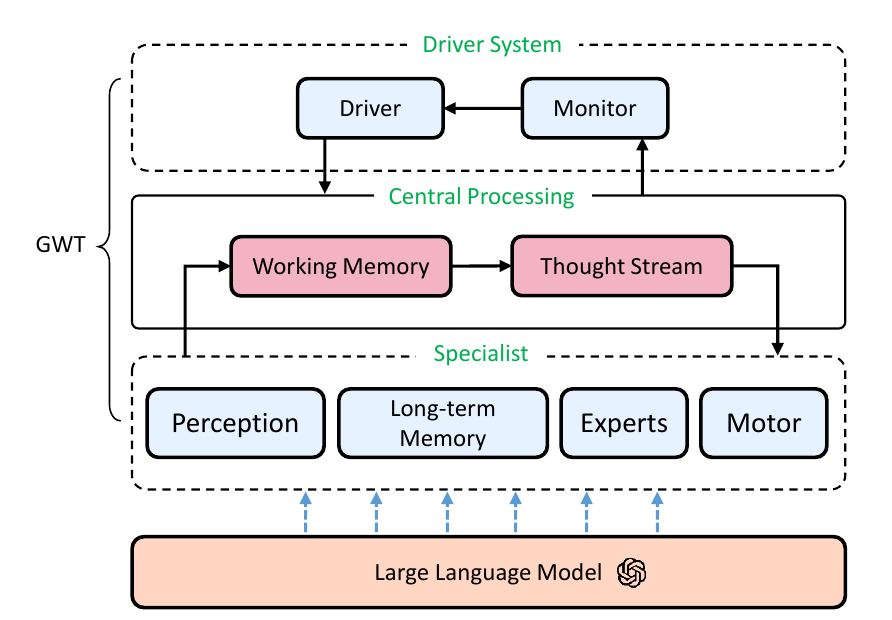}
    \caption{\label{fig:umm}
     The Architecture of UMM.  UMM is constructed around the Global Workspace Theory (GWT), which is organized into a hierarchical structure consisting of three distinct layers. The first layer is the Specialist module, which houses a variety of independent functional models. The second layer is the Central Processing module, which corresponds to the Global Workspace in GWT and governs the regulation and management of the Specialist module. The third layer is the Driver System, which corresponds to the Background Context in GWT and is responsible for dynamically adjusting the task objectives and information processing methodology. In addition, the UMM incorporates a language model that supports the implementation of various cognitive functions.
    }
\end{figure}

Specifically, the Unified Mind Model (UMM) is an intricate framework built upon the principles of Global Workspace Theory (GWT). GWT is structured hierarchically into three layers as showed in Figure \ref{fig:cog_architecture}. The foundational layer comprises various discrete execution units, namely specialist module, each dedicated to specific functions. The intermediate layer, termed the Global Workspace, corresponds to the Central Processing module and manages the lower-level functional modules. The top layer, known as the Background Context, aligns with the Driver System and guides the Global Workspace, dynamically adjusting task objectives and information processing methods. From a macro perspective, the GWT processes information in a specific sequence. Initially, it gathers relevant information for the current task, considering the background context. This information is then processed to generate decision data, which is broadcast to all experts in the specialist module for action execution. Upon task completion, these experts provide feedback, facilitating a continuous loop.  This sequence guides the system towards completing a specific task. Unlike other cognitive architectures, GWT can simulate macro-level collaborative computation across multiple brain regions.

Inspired by Global Workspace Theory \cite{baars2005global, baars1993cognitive}, UMM integrates large language models to support a wide range of human-like cognitive abilities, including perception, reasoning, planning, tool use, learning, memory, reflection, and motivation. We also introduce MindOS, an agent-building engine based on this cognitive architecture. MindOS allows users to easily create domain-specific autonomous agents by using free-form text to define attributes such as personality, motivation, background, domain knowledge, and tools, eliminating the need for programming skills. Our approach simplifies the creation of autonomous agents and serves as a valuable resource for both the research and application communities.

\section{The Theory Architecture of Unified Mind Model}
\label{sec:umm}
UMM is based on the Global Workspace Theory (GWT) \cite{baars1993cognitive,baars2005global} for its macro architecture. It integrates a large language model as a core element to enable advanced cognitive functions such as perception, learning, memory, action selection, tool use, reasoning, planning, reflection, and motivation. This section introduces the UMM architecture, focusing on its theoretical principles. Detailed implementation aspects are covered in the next section.

\subsection{Macro Architecture}
The human brain is a complex biological system characterized by distinct modular features in both structure and function. At a macroscopic level, it is divided into multiple functional modules that are interconnected morphologically and physiologically \cite{bertolero2015modular,meunier2010modular}. These modules collectively support human cognitive and behavioral capabilities. Traditional cognitive architectures often simplify the brain into five main functional areas: perception, motor (output), long-term memory, working memory, and motivation system \cite{laird2017standard,kotseruba2016review}. 

1. \textbf{Perception} processes various sensory stimuli from the external environment, such as visual, auditory, and tactile information, converting raw sensory data into neural signals and relaying them to other functional areas for further processing. 
2. \textbf{Motor Module} controls physical actions and behaviors, including muscle movements and speech. 
3. \textbf{Long-term Memory} involves storing and retrieving both long-term and short-term memories, where long-term memory retains past experiences and knowledge, and short-term memory temporarily holds recent information. 
4. \textbf{Working Memory} is considered the central processing system, consolidating data from various functional modules to facilitate cognitive activities. 
5. \textbf{Motivation System} drives task completion through methods such as desire fulfillment, rewards, or emotional inspiration.

These five functions collectively form a complete cognitive system through their interconnected interactions. Empirical studies have shown that these functional modules are sufficient to support a multitude of cognitive processes \cite{laird2017standard,laird2019soar,franklin2013lida,ritter2019act}.

In light of the substantial capabilities of large language models to facilitate a range of human-level cognitive functions, the UMM incorporates such a model to underpin the entire system's operations. The architecture of UMM is composed of four hierarchical modules as illustrated in Figure \ref{fig:umm}:
1. \textbf{Foundation Model Module}. This comprises various large language models.
2. \textbf{Specialist Module}. It includes multiple autonomous experts, each proficient in executing specific tasks independently. Additionally, this module encompasses perception, long-term memory, and motor functions, which operate independently under the governance of the central processing module.
3. \textbf{Central Processing Module}. Acting as the "central brain," it is responsible for the coordination and management of the entire system’s activities.
4. \textbf{Driver System}. This module modulates the central processing module's focus to enable the execution of specific tasks, thus playing a vital role in facilitating the autonomous behavior of the UMM.

The use of GWT as a macro architecture offers several benefits. First, specialist modules operating in parallel significantly boost information processing efficiency. This setup allows the system to flexibly and scalably add or remove experts in the specialist module. Second, the Central Processing module consolidates more task-relevant information from various specialist modules, facilitating more comprehensive and accurate decision-making and planning. Lastly, GWT allows for seamless integration of the language model's functionality into the sub-modules (see Section \ref{sec:umm-foundationmodel}), which considerably enhances overall system performance.

\subsection{Central Processing Module}
\label{sec:cpm}
The Central Processing module is tasked with gathering task-related data from the Specialist module to enable precise decision-making and planning for future steps. It comprises two sub-modules:

1. \textbf{Thought Stream.} It acts as an information processor, receiving aggregated task-related data and generating decisions and plans for subsequent steps.

2. \textbf{Work Memory.} It has two primary functions: First, it stores recently acquired data such as task-related memory, current perceptions, and global system states. Second, it aggregates this information into a basic processing unit termed "Thought," which supports the Thought Stream module in making accurate decisions and plans.

The Central Processing module operates with a workflow akin to the Global Workspace Theory (GWT). The process involves the following steps:

1. The Working Memory module collects task-relevant data from other modules, consolidating this information into a Thought.
2. The Thought Stream module then processes this Thought to generate decisions and plans for the next step.
3. These decisions and plans are disseminated to downstream Specialist Modules.
4. Specialist Modules engaged in the current task execute specific actions and generate results. These results are aggregated back by the Working Memory module for subsequent processing.

This cycle repeats continuously, enabling the system to complete the assigned tasks.

Throughout the execution, the Monitor module continually observes the Central Processing module, including both the Working Memory and Thought Stream. Its observations inform the Drive module, which adjusts the system's objectives, thus supporting dynamic control over the operation.

\subsection{Foundation Model}
\label{sec:umm-foundationmodel}
The Thought Stream module in UMM facilitates decision-making and planning, but achieving effective outcomes in these areas remains a significant challenge. Traditional cognitive architectures, such as SOAR \cite{laird2019soar}, ACT-R \cite{ritter2019act}, and LIDA \cite{franklin2013lida}, rely on handcrafted, symbolic procedural memory modules for task planning and decision-making. Upon receiving sensory input, these systems compare the input to existing procedural memories and select the most relevant one \cite{laird2017standard}. This procedural memory acts as a planner for subsequent decisions and actions.

However, the limitation of relying on handcrafted symbolic programs is that they are confined to predefined tasks within specific environments, which restricts their practical utility. An alternative approach is the implementation of a world model. For instance, LeCun et al. propose a cognitive architecture centered around a continually learning and adaptable world model \cite{lecun2022path}. This world model offers strong generalization capabilities, enabling the system to make accurate decisions and plans in open-domain environments. Therefore, we suggest it as a viable choice to support decision-making and planning within the Thought Stream module.

Fortunately, Large-scale language models like ChatGPT and GPT-4 exhibit remarkable task planning and reasoning capabilities. They have acquired extensive world knowledge, enabling them to adapt to diverse open-domain scenarios \cite{openai2023chat, openai2023gpt4}. We propose that these language models can act as world models. Consequently, we integrate the foundation model module into the UMM to enhance planning and decision-making.

Language models significantly enhance UMM operations by enabling effective module communication through free-text, thus maximizing system support capabilities. They serve as a world model, aiding accurate planning and decision-making. Language models handle information at varying granularities—words, sentences, or complex paragraphs—adding flexibility in information processing. They also learn extensive world knowledge, allowing for efficient handling of numerous System-1 tasks \cite{openai2023gpt4}, thus boosting UMM's task performance baseline. Moreover, studies show that language models excel when utilizing various tools \cite{schick2023toolformer, qin2023tool, shen2023hugginggpt, liang2023taskmatrix, chatplugin2023}. This mirrors the Central Processing model, coordinating multiple Specialist modules, and significantly enhancing UMM's coordination capabilities.

\subsection{Specialist Module}
In GWT, the Specialist Module comprises numerous unconscious experts capable of performing specific functions independently and in parallel. In the UMM, the IO module (Perception and Motor) and the Long-term Memory module are components of the Specialist Module.

\textbf{Expert.}
Experts are independent functional tools that can work in parallel. Experts provide UMM with the ability to interact with the digital world and perform complex tasks. 
In UMM, the range of experts is fairly broad, covering API calls, web search, program scripts, software, etc.

\textbf{Long-term Memory.}
Living organisms often store knowledge, experience, and task-specific information to enhance their survival capabilities in complex environments \cite{wood2012review}. Similarly, the Long-term Memory module in Unified Memory Models (UMM) functions analogously. Despite the challenges in developing human-like memory systems, various mechanisms have been explored. Neural network-based models for long-term memory, like Neural Turing Machines \cite{graves2014neural} and Key-Value Memory Networks \cite{miller2016key}, have been designed to theoretically store large volumes of information.

The rapid growth of internet information has greatly expanded the scope of human knowledge. Embedding this extensive knowledge directly into neural networks is both impractical and inefficient. As a result, traditional cognitive architectures typically store structured and unstructured data using databases or file systems, taking advantage of existing large-scale data storage solutions \cite{laird2019soar,ritter2019act}.

In practice, an effective long-term memory system should combine neural networks with file systems. Thus, UMM utilizes language models to store vast commonsense knowledge and employs file systems or databases to manage structured and unstructured domain-specific information. Furthermore, the internet can function as a long-term memory system within UMM, significantly boosting its knowledge base.

\textbf{IO.} In UMM, the I/O module comprises perceptual input and motor output. Perceptual input encompasses multiple modalities, such as images and text, while motor output is primarily conveyed through language. In practice, we enhance the I/O module with a language-based user interface to support more complex input and output forms. Details are provided in Section \ref{mindos:io}.

\subsection{Driver System}
Studying human motivation is a complex endeavor intersecting multiple disciplines, including neuroscience, psychology, and social sciences \cite{weiner2013human}. Motivation is the internal drive that compels individuals to act towards achieving specific goals. Basic biological motivations encompass food seeking, predator avoidance, and maintaining circadian rhythms such as sleep \cite{kesner2022seeking, sotelo2020sleep}. Typically, the intensity of motivation diminishes as needs are satisfied. Motivation acts as a driving force that helps organisms transition from unstable to stable states. It inspires, guides, and maintains certain behaviors, providing a robust sense of self-control \cite{reeve2018understanding}.

We propose a Driver System that allows the UMM to function autonomously. Due to the complexity of creating human-like motivation systems, we simplify the Driver System to an automated goal management system. The Driver System in UMM is composed of two key modules: (1) the Driver module, responsible for goal management and driving the entire system towards achieving set objectives; and (2) the Monitor module, which oversees the operations of the lower-level Central Processing module and reports task status to the Driver module. This structure enables the Driver module to dynamically adjust goals based on task progress, thus conferring a degree of autonomy to the UMM.

\section{MindOS}    
In this section, we introduce MindOS, an agent-building engine based on the UMM. MindOS enables users to rapidly create domain-specific autonomous agents equipped with high-level cognitive abilities. It leverages recent state-of-the-art algorithms and significantly enhances them.

\begin{figure}[!t] 
    \centering
    \includegraphics[width=1.0\columnwidth]{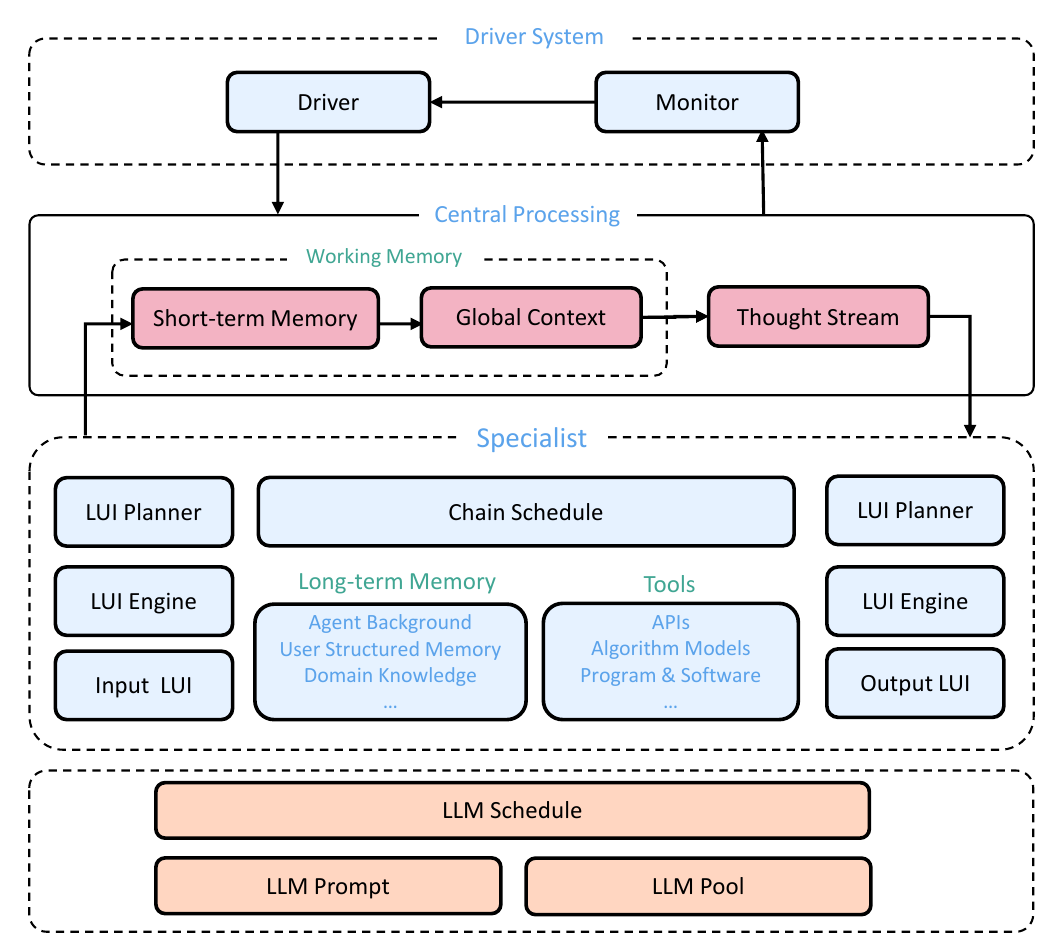}
    \caption{\label{fig:mindos}
        The Architecture of MindOS.
    }
\end{figure}

The macro architecture of MindOS, depicted in Figure \ref{fig:mindos}, features a four-tier hierarchical structure:

1. \textbf{Foundation Model Module.} This layer comprises various ChatGPT-like models. To effectively utilize these language models, we introduce three sub-modules for prompt selection and language model selection.
2. \textbf{Specialist Module.} This layer includes the IO module, Long-term Memory module, and various tools. MindOS tools encompass a wide range of operations in the digital realm, such as APIs, program scripts, web searches, and software operations.
3. \textbf{Central Processing Module.} Responsible for the coordination and control of the entire system, this module differs from UMM by splitting the Working Memory module into two distinct functional modules.
4. \textbf{Driver System.} Structurally identical to UMM, this layer ensures the overall functionality and autonomy of the system.

\subsection{Central Processing Module}
The Central Processing module is essential for managing and synchronizing the entire system. Supervised by the Driver System, it interacts with the Specialist module, both of which are supported by the language model.

\subsubsection{Thought Stream}
The Thought Stream module is a critical component for decision-making and planning in cognitive processing. It operates using a language model. The module receives two types of textual inputs: "Thought" from the Working Memory module and control signals from the Driver System. These inputs are processed by the language model, enabling decision generation and planning.

\subsubsection{Working Memory}
Working memory temporarily stores and manipulates information to aid in task completion. Despite its limited capacity and duration, it can store sufficient task-relevant information to support planning and decision-making, enhancing advanced cognitive functions \cite{cowan2014working}. In the UMM architecture, working memory includes two primary components: (1) Global Context, which integrates information from lower-level specialist modules and recent history to form a basic processing unit; and (2) Short-term Memory, which temporarily preserves historical information to enable coherent information processing.

\textbf{Global Context}
The Global Context module's primary function is to gather task-specific information from various components, including Working Memory, Long-term Memory, Driver System, and IO Module. It aggregates this information to form a fundamental processing unit, which is then forwarded to the Thought Stream module.

We term the basic information processing unit as a `Thought', a concept previously introduced in UMM. However, modeling a Thought remains an unresolved challenge. For humans, a Thought can manifest as a word, sentence, sound, image \cite{smallwood2003task}, or contextual mental representation such as time, location, or scene \cite{kane2007whom,kvavilashvili2004out}. Despite the varying granularity of this information, the human brain processes it efficiently. To date, there is no standardized definition for a `Thought' within neuroscience and psychology.

In MindOS, we represent the 'Thought' using a structured prompt, leveraging the language model's prompt-based framework. This method allows the Thought Stream model to fully utilize the language model's capabilities. This approach provides two main benefits. First, the language model can handle highly unstructured textual inputs, making the prompt's length, granularity, and format adaptable. This flexibility is crucial for the Thought Stream module. Second, the language model's robust task planning abilities \cite{xie2023translating, song2022llm, huang2022language, wang2023describe} significantly enhance the functionality of the Thought Stream module.

Consequently, we model the process of gathering task-related information within the Global Context as the creation of a structured prompt, referred to as "Thought." The Thought is organized into the following components:
\begin{itemize}[leftmargin=2em]
\item Instructions. Task instructions sent from the Driver system.
\item Dialog Context. The user's conversation history, stored primarily in Working Memory.
\item Perception Information Sensory. The data obtained from the environment and processed through the Language User Interface (see Section \ref{mindos:io}), then represented in text format.
\item User Profile. This includes the user's preferences, habits, and other characteristics in specific scenarios, aiming to help the agent better understand the user.
\item Agent Profile. This includes background information about the agent.
\item Related Memory. Information retrieved from the Long-term Memory module.
\item History. Incorporates Agent Action, User Action, and other operational history, mainly stored in Working Memory.
\item Date. The current time.

This structure ensures a comprehensive and organized approach to task-related information gathering, enhancing the agent's ability to generate relevant and accurate responses.
\end{itemize}

\textbf{Short-term Memory}

Short-term Memory temporarily stores task-related history, status information, dialogue context, perceptual data, relevant memories, and historical actions during interaction sessions. This information is accessed by the Global Context module, encapsulated into Thoughts, and then sent to the Thought Stream module for decision-making and planning. Specifically, Short-term Memory primarily stores the following types of information:
\begin{itemize}[leftmargin=2em]
\item Agent Action. This pertains to the action history carried out by the Agent, such as invoking the web search API to retrieve relevant data or querying task-related information from the Long-term Memory module.
\item User Action. When a user interacts within a specific environment, their actions are recorded by the embedded agent. For instance, on a shopping site, if a user clicks a button or browses certain products, the agent records these interactions. This record assists the agent in understanding the user's behavior better.
\item Mind \& User Conversation. The history of conversations between the agent and the user during a session.
\item Scene Information. The context of the agent's environment and the current time.
\end{itemize}

\subsection{Foundation Model}
In MindOS, the Foundation Model Module orchestrates various language models to support the entire system's operation. This module is divided into three sub-layers:

1. LLM Pool Layer. Contains a diverse collection of ChatGPT-like language models.
2. Prompt Layer. Offers numerous prompts for each language model, including instruction prompts, few-shot demonstrations, and various prompt templates. These preset prompts enhance the system's ability to leverage the language models effectively, ensuring more stable support for higher-level modules.
3. LLM Schedule Layer. Selects the appropriate language model and corresponding prompts for specific tasks.
This structured approach ensures the efficient and reliable operation of the system's language processing capabilities.

\subsection{Specialist Module}
The Specialist Module comprises numerous independent functional experts. To avoid confusion, we refer to these experts as tools, akin to their usage in Agent communities \cite{schick2023toolformer,qin2023tool,shen2023hugginggpt,liang2023taskmatrix,chatplugin2023}. These tools range from simple APIs to complex program scripts and algorithm models. Given that the agent primarily operates in a digital environment, it can utilize web search tools, computer software, and even other agents created by MindOS. These tools are accessible via a unified interface on MindOS, enabling Thought Stream to orchestrate their use for completing complex tasks. MindOS comes with various built-in tools, including image creation, web searching, and product inquiries. Users can select tools that best meet their needs without any programming effort. It is important to highlight that in MindOS, the IO module and the Long-term Memory module are also considered part of the Specialist Module. For clarity, we discuss these modules separately in other sections.

Leveraging the advanced code comprehension capabilities of contemporary language models \cite{openai2023gpt4,openai2023chat}, MindOS supports seamless integration of simple tool configurations. This integration necessitates defining only the program interface and function description for each tool. MindOS can incorporate new tools efficiently through two methods: (1) Direct import of YML files in the OpenAPI format, including configurations for interfaces, function descriptions, and related details. (2) Utilizing MindOS's intuitive interactive interface, where users can add new tools by providing the function description, API addresses, and input/output parameters

According to the theoretical architecture of GWT, decision and planning information generated by the global workspace is broadcast to all specialist modules. Upon receiving the broadcast, these modules process the information and send it back to the central processing module to support subsequent operations \cite{baars1993cognitive,baars2005global}. In contrast, MindOS uses a paradigm where tools feature clear interfaces, eliminating the need for such broadcasts. Instead, information is parsed and sent directly to the relevant tools, markedly enhancing system efficiency. This improvement is attributed to the robust semantic understanding and tool utilization capabilities of the language model, enabling MindOS to dynamically select the necessary tools \cite{chatplugin2023, shen2023hugginggpt}.

In addition, the language model demonstrates a robust capacity to utilize a chain of tools for executing complex tasks \cite{xie2023translating,vemprala2023chatgpt,liang2023taskmatrix,shen2023hugginggpt,chatplugin2023}. Leveraging these capabilities, MindOS efficiently orchestrates the use of various tools within a defined workflow. This is facilitated by scheduling tool usage in a predetermined sequence, thus eliminating the need for human intervention. The Thought Stream module generates a detailed plan, specifying the necessary tools and their sequence to complete a given task. The Chain Schedule module subsequently receives this plan and manages the execution of the tools.

\subsection{Long-term Memory}
In MindOS, the Long-term Memory module functions as a sophisticated file system, storing information in various formats, including free text, files, and structured data. It supports diverse operations such as querying, adding, and deleting data. Furthermore, MindOS integrates with web search tools, enabling the system to leverage extensive Internet data. The Long-term Memory module comprises the following memory types:
\begin{itemize}[leftmargin=2em]
    \item  Agent Profile. The profile information that users have set for the created agent, as well as the events, dialogues, and scenes that the agent encountered during the runtime.
    \item  User profile. The user's basic information, behavior preferences, habits, etc.
    \item  User structured memory. The experience of users,including dialog, behavior,event,etc.
    \item  Domain knowledge. The expert knowledge in a specific field.
    \item  Tools. Tool's configuration information, including tool descriptions, interface definitions, so that the agent ``knows'' what tools it can use.
\end{itemize}

In MindOS, adding new memories to the Long-term Memory module is straightforward:
1. To include profile information for the agent, simply upload a description file, which will be stored in free text format.
2. The user profile and structured memory are saved asynchronously in free text and key-value formats during user interaction (with user authorization).
3. The user's conversation history is stored in free text format in the database to help the agent better understand the user (with user authorization).
4. For domain knowledge, upload files in free-form or structured text. MindOS will automatically analyze and create an embedding index for the file. It also supports structured data, given that a detailed interface is defined during the upload.
5. Information from the Internet is retrieved via APIs, with critical information also being cached.

In addition, the language model, with its extensive world knowledge, can function as a memory system. Within MindOS, the language model is pivotal in information processing. When a user's request surpasses the agent's Long-term Memory, the agent leverages the language model's world knowledge to make informed decisions. This augmentation improves the system's adaptability to new situations while preserving its core functionality.

\subsection{IO}
\label{mindos:io}

Digital environments offer richer interaction possibilities compared to physical settings, such as clicking buttons and selecting options. Basic inputs like text and images are not sufficient to capture this complexity. In MindOS, we utilize a Language User Interface (LUI) for I/O interactions. The LUI supports both simple inputs—like sentences, speech, images, and files—and complex actions, such as clicking buttons and navigating menus. The LUI Planner is a critical component with dual functions. First, it parses input information and converts it into language-based data, enhancing system understanding. Second, it decodes system outputs and plans a UI layout to optimize information display and user interaction. This blend of input forms enhances user-agent interactions, resulting in a more efficient and satisfactory user experience.

\subsection{Driver System}
\label{sec:drive_system}
In MindOS, the Driver System regulates information processing by dynamically adjusting goals to enable autonomous task performance. This is achieved through two modules: Driver Module and Monitor Module.

\subsubsection{Driver Module}
In MindOS, the Thought Stream module leverages a language model to process information, transforming it into Thought prompts. Consequently, the Driver module can regulate the Thought Stream by adding specific instruction prompts to achieve certain objectives.

The Driver module manages three types of behavior drivers:
1. Long-term drives: These encompass an agent's fundamental background elements, including initial motivation and persona, set during the construction phase. For instance, "As a teacher, your objective is to excel in your teaching duties." This drive continually influences an agent's behavior throughout its lifecycle.
   
2. Short-term drives: Users may require agents to complete particular tasks during interactions. The fulfillment of these tasks represents short-term motivation. User goals are often decomposed into sub-tasks during system operation, each associated with a short-term drive.
   
3. Reactive motivations: These trigger specific system actions upon receiving pre-set inputs. In such scenarios, the system can execute predefined responses without engaging in Thought Stream processing.

This structure allows for a dynamic and responsive system capable of detailed and nuanced behavior management.

\subsubsection{Monitor Module}
To dynamically adjust the goal, a Monitor module is essential to track the current execution status. At each time step, this module observes the inputs and outputs of the Thought Stream to ascertain the system's status.

The Monitor operates in two modes:
1. For long-term and short-term drives, where the Drive model mainly uses prompt injection to adjust the Thought Stream module’s behavior. The Monitor observes three key pieces of information: first, the termination condition, to determine if the current task is complete; second, the number of steps taken, and if it exceeds a predefined threshold, the system halts the task to avoid deadlock; third, whether a sub-task needs execution, which helps in setting sub-goals for the Drive module.

2. The Monitor also stores pre-set Trigger records. When the Thought constructed by the Global Context matches a pre-set Trigger, the Monitor immediately adjusts the system to output the corresponding response, bypassing the Thought Stream module.

This structured monitoring ensures dynamic goal adjustment and efficient task management.

\subsection{Three Types of Information Processing Mode}

The principal role of the Thought Stream module is to receive inputs and generate corresponding plans and decisions. Throughout this process, its input and output are continuously monitored by the Monitor module. The Driver model subsequently adjusts the goals and modifies the operational mode of the Thought Stream according to the current task state. Despite this straightforward process, MindOS can carry out three types of task implementation pipelines through the synergy of the Driver System and Central Processing modules.

\begin{figure}[!t] 
    \centering
    \includegraphics[width=1.0\columnwidth]{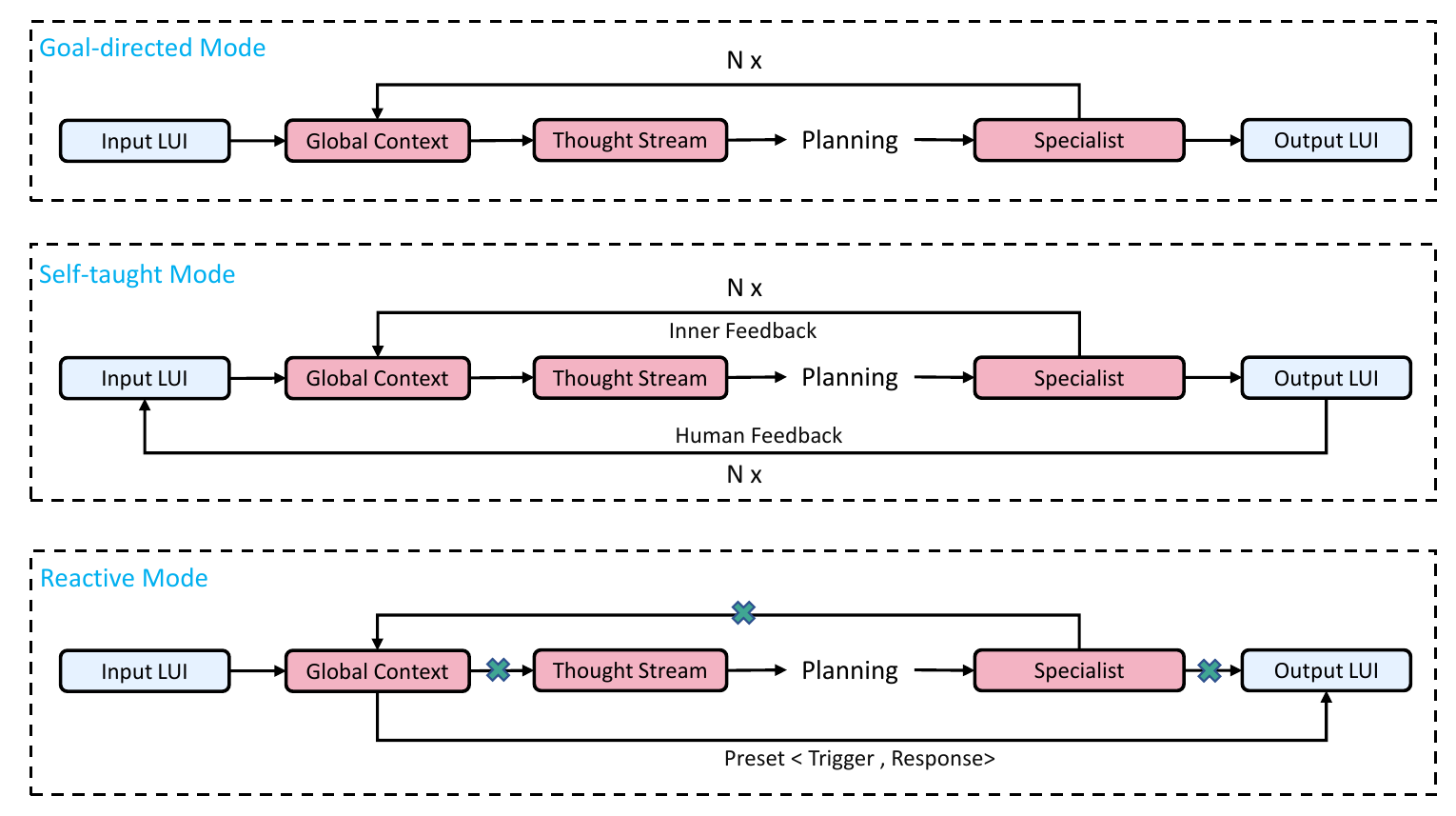}
    \caption{\label{fig:pipeline}
        Information Processing Mode.
    }
\end{figure}

\paragraph{Goal-directed Mode.} It is designed to solve specific tasks, as illustrated in Figure \ref{fig:pipeline}. This mode handles two types of tasks: short-round tasks and workflow tasks.
For short-round tasks such as dialogue, question answering, or information retrieval, the Thought Stream first generates a plan to retrieve task-related knowledge or query information using web search tools. This information is then aggregated to form the next step in the Thought process. Once sufficient information is gathered, the Thought Stream generates the final answer, which is sent to the LUI for user output.

For workflow tasks that require multi-step completion, the Thought Stream initially generates a plan to divide the goal into sub-goals. It then iteratively executes the "Thought Stream -> planning -> Specialist" procedure to accomplish all sub-goals progressively. After completion, the results are sent to the LUI for user output. This mode incorporates "Inner thinking" processes and supports the completion of complex tasks.

\paragraph{Self-taught Mode.} This mode autonomously learns new knowledge and discovers new solutions to given problems or tasks. Through iterative planning, output generation, and feedback reception, it eventually identifies the answer. This approach allows the incorporation of novel tools, the discovery of alternative solutions, and the handling of open-domain problems. Unlike the goal-directed mode, the self-taught mode can receive feedback from both internal tools and humans. It can store learned workflows and tools, thus avoiding redundant procedures in future instances. By retrieving these stored workflows and tools, it can improve the performance of similar tasks.

\paragraph{Reactive Mode.} This mode provides an immediate response based on a pre-set trigger. This type of behavior is commonly used in the design of intelligent systems for specific scenarios. For example, if a user shuts down an agent, it may trigger a query response to secure the system. The support of MindOS allows it to quickly adapt to specific environments and tasks by responding to input signals, which improves the flexibility and robustness of the system. In addition, it can support event and timed triggering, which greatly enhances the usability of the system. In Mindos, the user can setting ``trigger-response'' pairs for agent, this enable the agent generate the expect behavior when receive the corresponding trigger.

\subsection{Learning}
Learning is crucial for all living organisms as it enables the acquisition of new knowledge and skills to better adapt to complex environments. Agents developed by MindOS primarily inhabit the digital world, where their learning objectives include understanding human behavior, gaining expert knowledge, developing new skills, and mastering task-solving workflows. We propose three main learning methods for MindOS, each designed to enhance the system from a different perspective.

\subsubsection{Modular Learning}
MindOS features a modular system architecture that simplifies the integration and disassembly of various modules. Consequently, acquiring a new skill is akin to adding a new tool to MindOS. This tool can range from an API, an algorithm model, a program script, to an application interface. This design allows MindOS to rapidly expand and enhance its capabilities.

\subsubsection{Language Model Learning}
Currently, the most effective method for augmenting language models employs the same training strategy as ChatGPT and GPT-4, which utilize a combination of supervised learning and RLHF \cite{ouyang2022training}. Given the rapid iterations of large-scale language models, MindOS adopts two optimization strategies: (1) directly replacing the language model with a more advanced one, and (2) continually fine-tuning the existing language model. By interacting with users, MindOS can accumulate a substantial amount of high-quality data, which can then be used for further fine-tuning the language model.

\subsubsection{Auto Prompt Learning} 
In MindOS, the fundamental information processing unit is a structured prompt. Because language models are sensitive to prompt changes, enhancing the prompt quality can boost the model's performance \cite{kojima2022large}. Various modules within MindOS, such as Though Stream, Drive System, Long-term Memory, and LUI Planner, heavily rely on the language model. Consequently, refining prompts is crucial for the overall performance of the MindOS system. We pre-configure handcrafted optimal prompts for different language models, which can be refined offline. Given the diverse scenarios in which agents operate, it is advisable to explore automated learning techniques for improving prompt generation. Nonetheless, current auto-prompt approaches are limited to specific tasks and simple prompt structures \cite{shao2023synthetic,zhang2022automatic,deng2022rlprompt}. MindOS, with its complex prompt structure, necessitates the development of new methods for online prompt optimization to achieve more efficient and stable learning outcomes.

\subsubsection{Workflow Learning} 
In MindOS, workflows are learned through two approaches. First, users can manually configure a specific workflow for a task, allowing MindOS to learn directly. Second, workflows can be learned via the self-taught mode, where the agent interacts with users or tools within the Specialist module to discover a suitable workflow for solving the given problem.

MindOS then retains successful workflow trajectories, aiding accurate decision-making in similar future tasks. Additionally, the acquired workflows can be used to fine-tune the language model directly, enabling it to make precise decisions without relying on external records. This process is akin to transferring experience from System 2 to System 1, thereby improving task-solving efficiency \cite{frankish2010dual}.

\subsection{Discussion}
\textbf{What is the difference between MindOS and existing works?}
Numerous autonomous agents have been proposed around large language models, such as AutoGPT \cite{autogpt2023}, HuggingGPT \cite{shen2023hugginggpt}, and LangChain agent \cite{langchain2023agent}. These agents leverage language models to autonomously solve problems using various tools. Based on given instructions, they can automatically plan executable sub-steps and proceed to execute tasks accordingly.

MindOS distinguishes itself in two key ways. First, while these agents are essentially autonomous task-solving systems, the agent created by MindOS is rooted in a comprehensive cognitive architecture. This encompasses a broader range of high-level cognitive capabilities, including motivation, introspection, learning, and memory. Second, MindOS serves as a more powerful agent-building engine, allowing users to quickly create domain-specific autonomous agents without any programming effort.

\textbf{Can LLM be regarded as world model?}
Cognitive architecture, dating back to the 1950s, aimed to develop general-purpose AI by simulating human thought processes. Despite decades of work, the complexity of human cognition has prevented the community from agreeing on a unified definition of cognitive abilities. As a result, existing cognitive architectures vary greatly in structure and function.

\begin{figure}[!t] 
    \centering
    \includegraphics[width=1.0\columnwidth]{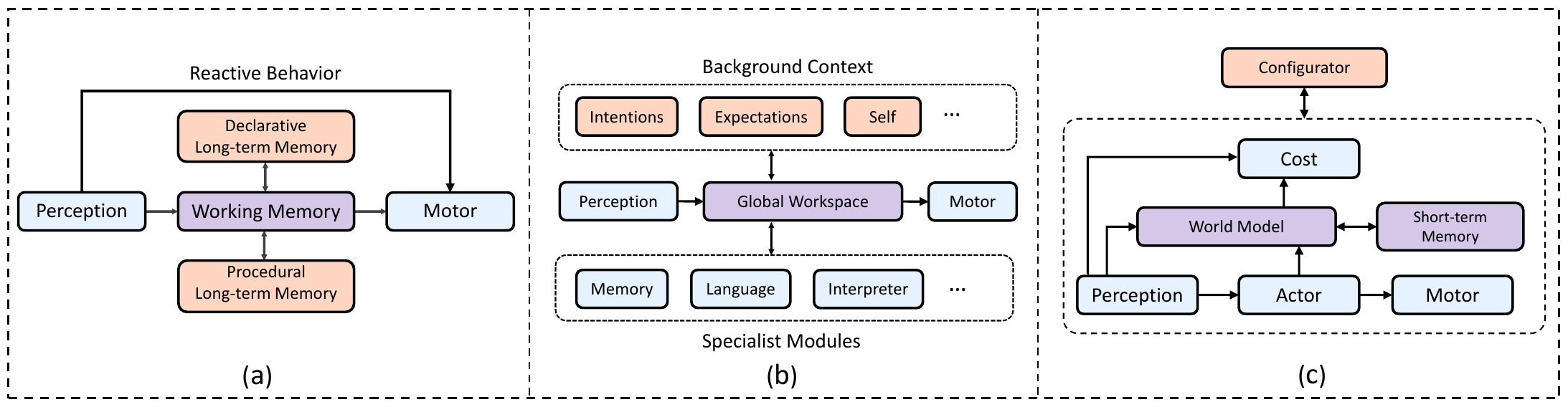}
    \caption{\label{fig:cog_architecture}
        Three types of Architecture. (a) A standard cognitive architecture \cite{laird2017standard}. (b) Architecture of Global Workspace Theory \cite{baars1993cognitive}. (c) Lecun's Architecture \cite{lecun2022path}. 
    }
\end{figure}

From a macro-architecture perspective, there are three types of cognitive architectures as shown in Figure \ref{fig:cog_architecture}:
1. Procedural Memory-Centered Architecture: This traditional cognitive architecture relies on handcrafted procedural memory to execute specific tasks \cite{ritter2019act,laird2019soar,kotseruba2016review}. It exhibits weak generalization and is not suitable for open-ended scenarios.

2. GWT-Based Cognitive Architecture: These architectures operate similarly to the GWT at a macro level. They typically require a procedural memory module or an equivalent function module to plan and decide on the next action \cite{franklin2013lida,arrabales2009cera}.

3. World Modeling-Centered Architecture: The most advanced cognitive architecture, as exemplified by LeCun's autonomous architecture \cite{lecun2022path}, centers around a world model. This model enables the system to learn new skills, perform complex planning, make decisions, and predict outcomes, leading to accurate and reasonable actions. The overall goal is to minimize intrinsic cost, driving the system to interact autonomously with the external environment. World models overcome the limitations of traditional procedural memory modules, providing flexible learning of new skills and strong generalization capabilities.

In fact, the language model shares similarly with the world model or procedural memory, since all of them can make complex planning and reasoning. This observation leads to the plausible that use language model to replace the position of world model in traditional cognitive architecture. The incorporation of a language model to UMM is a fundamental rationale behind this observation. In contrast to existing cognitive architectures, MindOS is comprehensively supported by currently powerful LLMs. As expounded in Section \ref{sec:umm-foundationmodel}, this feature confers numerous advantages to the overall system.

\textbf{How to build rich inner world for autonomous agents?}
Recent psychological studies have identified two distinct thought patterns in the human brain. The first is Goal-directed Thought, a mode activated during specific task execution. In this mode, individuals adhere to a specific procedure \cite{botvinick2014computational,evans2008dual,miller2000prefontral}, which limits creativity \cite{dosenbach2006core,hazeltine2016understanding}. The second is Spontaneous Thought, an endogenous information processing mode occupying 30-50\% of waking hours \cite{kane2007whom,klinger1987dimensions}. Psychological phenomena such as daydreaming, mind-wandering, and creative thinking \cite{dijksterhuis2006creativity,christoff2011role} are categorized under Spontaneous Thought. The human inner world is primarily composed of these two thought patterns.

Spontaneous Thought plays several crucial roles in the human brain. Studies estimate that approximately 20\% of Spontaneous Thought involves future prediction and planning \cite{kane2007whom,smallwood2009your}, indicating that this cognitive process functions similarly to a world model. Additionally, Spontaneous Thought often retrieves and replays segments of past memory, aiding in the generalization of past experiences \cite{fields2002we,buckner2008brain,stawarczyk2011mind}. This suggests that Spontaneous Thought helps optimize a "world model" within the brain.

Research by Dijksterhuis et al. shows that individuals who engage in more mind wandering during specific tasks exhibit greater creativity \cite{dijksterhuis2006creativity}. For example, taking breaks, going for walks, daydreaming, or sleeping when faced with challenging tasks often leads to better solutions. This phenomenon occurs because individuals more readily switch to the Spontaneous Thought mode during these activities. Consequently, Spontaneous Thought facilitates creative thinking \cite{camfield2005neurobiology,singer2009researching}.

Furthermore, Christoff et al. observe that Spontaneous Thought commonly occupies the mind during rest, with brain activity resembling that of memory consolidation processes occurring during sleep \cite{christoff2011role}. This indicates that Spontaneous Thought assists in memory consolidation. Moreover, Spontaneous Thought can integrate isolated past experiences into a coherent and meaningful biographical structure. This allows individuals to perceive their past coherently and develop a sense of self \cite{christoff2011role}, a capability currently absent in artificial cognitive systems.

In the wake hour, these two modes switch frequently, forming a thought stream\cite{christoff2012undirected,christoff2004neural}. Spontaneous Thought and Goal-directed thought represent the processes of Exploration and Exploitation, respectively. 

However, the existing cognitive architectures, including our MindOS, are built around the Goal-directed Thought mode, which only enables systems to autonomously execute specific tasks, instead of owning a rich inner world in the mind.  In practice, MindOS normally retrieve task-related memory
and tools to help solving a task with many ``inner thinking'' steps. This procedure works similarly to CoT \cite{wei2022chain} or ReAct \cite{yao2022react}. However, how to model spontaneous thought is still an open problem. In the future, we will make more exploration for building the ``inner world'' of the agent.

\textbf{What's the limitation of Driver System?}
In theory, motivation can be categorized into two types:
1. Innate Motivation: This is an inherent drive present from birth, stemming from an organism's internal homeostatic system \cite{emde1988development}. It remains constant throughout the organism's development.

2. Acquired Motivation: Also known as indirect motivation, this type develops through environmental learning \cite{solomon1980opponent}. It is closely linked to innate motivation. For instance, humans often strive for money because societal interactions teach them that money fulfills fundamental needs, such as purchasing food to alleviate hunger. Hence, there is a strong correlation between socially learned motivations, the pursuit of money, and the innate drive to mitigate hunger.

In UMM, we use a language model-based driving system to model motivation. By injecting a prompt to the Thought Stream module, it can directly adjust the goal of the entire system.  This approach integrates well with the language model and is able to perform some autonomous behavior. However, it still far away from the human-like motivation systems.

\textbf{How to process other modalities in MindOS?}
In MindOS, we deploy a large language model to support the system's overall operation, which currently processes only text-based inputs. To enable the system to recognize other modalities, we employ various tools to convert these modalities into textual information. This method is similar to techniques like HuggingGPT \cite{shen2023hugginggpt} and ChatGPT Plugins \cite{chatplugin2023}. 

However, converting other modalities to text often results in information loss and reduced efficacy in handling multi-modal tasks. A potential solution is to directly use language-centered visual-language models, such as Flamingo \cite{alayrac2022flamingo}, or the multi-modal version of GPT-4 \cite{openai2023gpt4}, as the foundational model for MindOS. We plan to conduct further research to explore the feasibility of adopting these models.

\textbf{What is a better learning mechanism?}
In this study, we explore the development of four distinct learning modes within MindOS: learning tools, learning language model, learning prompt, and learning workflow.  Specifically, MindOS focuses on optimizing task execution and agent performance, thereby restricting its ability to learn flexibly across diverse domains as humans do. Additionally, the learning process in MindOS is discrete, hindering its capacity for continuous learning over time, which is essential for forming a comprehensive autobiographical memory. Further research and development are needed to enhance the learning capabilities of MindOS.

\textbf{What are the future challenges of cognitive architecture?}
Drawing inspiration from GWT, the UMM integrates language models to boost functionality. Though the current macro architecture of UMM effectively supports autonomous agents with human-level abilities, it requires more advanced designs for complex cognitive tasks like emotion, motivation, long-term learning, and memory. Developing these designs presents significant challenges in both macro architecture and underlying algorithms. Our ongoing research focuses on optimizing these components for continuous improvement. We urge researchers to prioritize macro cognitive architecture studies to advance the development of the Human-level Agent."

\section{Conclusion}
In this paper, we utilize the Global Workspace Theory (GWT), a renowned cognitive framework, to propose the Unified Mind Model (UMM) for building intelligent agents. By situating Large Language Models (LLMs) within the GWT, we demonstrate their potential to replicate human cognitive processes. LLMs, which proficiently handle vast linguistic data and generate adaptive responses, are prime candidates for simulating complex cognitive functions. We also introduce MindOS, an agent development platform grounded in the UMM framework, which integrates various functional modules to provide robust structural support for these agents. We explore the implementation principles, operational mechanisms, and performance of MindOS across multiple application scenarios. In conclusion, we aim for this work to inspire the academic community to consider new perspectives on the use of LLMs in advancing cognitive science and AI. We are hopeful that through interdisciplinary collaboration and continuous innovation, we can create more intelligent and efficient human-level agent systems, thus offering significant benefits to society.

\bibliographystyle{plain}
\bibliography{reference}

\begin{thebibliography}{10}

\bibitem{alayrac2022flamingo}
Jean-Baptiste Alayrac, Jeff Donahue, Pauline Luc, Antoine Miech, Iain Barr, Yana Hasson, Karel Lenc, Arthur Mensch, Katherine Millican, Malcolm Reynolds, et~al.
\newblock Flamingo: a visual language model for few-shot learning.
\newblock {\em Advances in Neural Information Processing Systems}, 35:23716--23736, 2022.

\bibitem{arrabales2009cera}
Raul Arrabales, Agapito Ledezma, and Araceli Sanchis.
\newblock Cera-cranium: A test bed for machine consciousness research.
\newblock In {\em International workshop on machine consciousness}, volume 272, 2009.

\bibitem{baars1993cognitive}
Bernard~J Baars.
\newblock {\em A cognitive theory of consciousness}.
\newblock Cambridge University Press, 1993.

\bibitem{baars2005global}
Bernard~J Baars.
\newblock Global workspace theory of consciousness: toward a cognitive neuroscience of human experience.
\newblock {\em Progress in brain research}, 150:45--53, 2005.

\bibitem{bertolero2015modular}
Maxwell~A Bertolero, BT~Thomas Yeo, and Mark D’Esposito.
\newblock The modular and integrative functional architecture of the human brain.
\newblock {\em Proceedings of the National Academy of Sciences}, 112(49):E6798--E6807, 2015.

\bibitem{botvinick2014computational}
Matthew~M Botvinick and Jonathan~D Cohen.
\newblock The computational and neural basis of cognitive control: charted territory and new frontiers.
\newblock {\em Cognitive science}, 38(6):1249--1285, 2014.

\bibitem{bubeck2023sparks}
S{\'e}bastien Bubeck, Varun Chandrasekaran, Ronen Eldan, Johannes Gehrke, Eric Horvitz, Ece Kamar, Peter Lee, Yin~Tat Lee, Yuanzhi Li, Scott Lundberg, et~al.
\newblock Sparks of artificial general intelligence: Early experiments with {GPT-4}.
\newblock {\em arXiv preprint arXiv:2303.12712}, 2023.

\bibitem{buckner2008brain}
Randy~L Buckner, Jessica~R Andrews-Hanna, and Daniel~L Schacter.
\newblock The brain's default network: anatomy, function, and relevance to disease.
\newblock {\em Annals of the new York Academy of Sciences}, 1124(1):1--38, 2008.

\bibitem{camfield2005neurobiology}
David Camfield.
\newblock Neurobiology of creativity.
\newblock {\em Neurobiology of exceptionality}, pages 53--72, 2005.

\bibitem{chen2023teaching}
Xinyun Chen, Maxwell Lin, Nathanael Sch{\"a}rli, and Denny Zhou.
\newblock Teaching large language models to self-debug.
\newblock {\em arXiv preprint arXiv:2304.05128}, 2023.

\bibitem{christoff2012undirected}
Kalina Christoff.
\newblock Undirected thought: neural determinants and correlates.
\newblock {\em Brain research}, 1428:51--59, 2012.

\bibitem{christoff2011role}
Kalina Christoff, Alan Gordon, and Rachelle Smith.
\newblock The role of spontaneous thought in human cognition.
\newblock In {\em Neuroscience of decision making}, pages 271--296. Psychology Press, 2011.

\bibitem{christoff2004neural}
Kalina Christoff, Justin~M Ream, and John~DE Gabrieli.
\newblock Neural basis of spontaneous thought processes.
\newblock {\em Cortex}, 40(4-5):623--630, 2004.

\bibitem{cowan2014working}
Nelson Cowan.
\newblock Working memory underpins cognitive development, learning, and education.
\newblock {\em Educational psychology review}, 26:197--223, 2014.

\bibitem{deng2022rlprompt}
Mingkai Deng, Jianyu Wang, Cheng-Ping Hsieh, Yihan Wang, Han Guo, Tianmin Shu, Meng Song, Eric~P Xing, and Zhiting Hu.
\newblock Rlprompt: Optimizing discrete text prompts with reinforcement learning.
\newblock {\em arXiv preprint arXiv:2205.12548}, 2022.

\bibitem{dijksterhuis2006creativity}
Ap~Dijksterhuis and Teun Meurs.
\newblock Where creativity resides: The generative power of unconscious thought.
\newblock {\em Consciousness and cognition}, 15(1):135--146, 2006.

\bibitem{dosenbach2006core}
Nico~UF Dosenbach, Kristina~M Visscher, Erica~D Palmer, Francis~M Miezin, Kristin~K Wenger, Hyunseon~C Kang, E~Darcy Burgund, Ansley~L Grimes, Bradley~L Schlaggar, and Steven~E Petersen.
\newblock A core system for the implementation of task sets.
\newblock {\em Neuron}, 50(5):799--812, 2006.

\bibitem{emde1988development}
Robert~N Emde.
\newblock Development terminable and interminable. i. innate and motivational factors from infancy.
\newblock {\em The International Journal of Psycho-Analysis}, 69:23, 1988.

\bibitem{evans2008dual}
Jonathan St~BT Evans.
\newblock Dual-processing accounts of reasoning, judgment, and social cognition.
\newblock {\em Annu. Rev. Psychol.}, 59:255--278, 2008.

\bibitem{fields2002we}
Chris Fields.
\newblock Why do we talk to ourselves?
\newblock {\em Journal of Experimental \& Theoretical Artificial Intelligence}, 14(4):255--272, 2002.

\bibitem{frankish2010dual}
Keith Frankish.
\newblock Dual-process and dual-system theories of reasoning.
\newblock {\em Philosophy Compass}, 5(10):914--926, 2010.

\bibitem{franklin2013lida}
Stan Franklin, Tamas Madl, Sidney D’mello, and Javier Snaider.
\newblock Lida: A systems-level architecture for cognition, emotion, and learning.
\newblock {\em IEEE Transactions on Autonomous Mental Development}, 6(1):19--41, 2013.

\bibitem{graves2014neural}
Alex Graves, Greg Wayne, and Ivo Danihelka.
\newblock Neural turing machines.
\newblock {\em arXiv preprint arXiv:1410.5401}, 2014.

\bibitem{hazeltine2016understanding}
Eliot Hazeltine and Eric~H Schumacher.
\newblock Understanding central processes: The case against simple stimulus-response associations and for complex task representation.
\newblock In {\em Psychology of learning and motivation}, volume~64, pages 195--245. Elsevier, 2016.

\bibitem{huang2022language}
Wenlong Huang, Pieter Abbeel, Deepak Pathak, and Igor Mordatch.
\newblock Language models as zero-shot planners: Extracting actionable knowledge for embodied agents.
\newblock In {\em International Conference on Machine Learning}, pages 9118--9147. PMLR, 2022.

\bibitem{langchain2023agent}
hwchase17.
\newblock Langchain.
\newblock {\em https://github.com/hwchase17/langchain}, 2023.

\bibitem{kane2007whom}
Michael~J Kane, Leslie~H Brown, Jennifer~C McVay, Paul~J Silvia, Inez Myin-Germeys, and Thomas~R Kwapil.
\newblock For whom the mind wanders, and when: An experience-sampling study of working memory and executive control in daily life.
\newblock {\em Psychological science}, 18(7):614--621, 2007.

\bibitem{kesner2022seeking}
Andrew~J Kesner, Coleman~B Calva, and Satoshi Ikemoto.
\newblock Seeking motivation and reward: Roles of dopamine, hippocampus, and supramammillo-septal pathway.
\newblock {\em Progress in neurobiology}, 212:102252, 2022.

\bibitem{klinger1987dimensions}
Eric Klinger and W~Miles Cox.
\newblock Dimensions of thought flow in everyday life.
\newblock {\em Imagination, Cognition and Personality}, 7(2):105--128, 1987.

\bibitem{kojima2022large}
Takeshi Kojima, Shixiang~Shane Gu, Machel Reid, Yutaka Matsuo, and Yusuke Iwasawa.
\newblock Large language models are zero-shot reasoners.
\newblock {\em arXiv preprint arXiv:2205.11916}, 2022.

\bibitem{kotseruba2016review}
Iuliia Kotseruba and John~K Tsotsos.
\newblock A review of 40 years of cognitive architecture research: Core cognitive abilities and practical applications.
\newblock {\em arXiv preprint arXiv:1610.08602}, 2016.

\bibitem{kvavilashvili2004out}
Lia Kvavilashvili and George Mandler.
\newblock Out of one’s mind: A study of involuntary semantic memories.
\newblock {\em Cognitive psychology}, 48(1):47--94, 2004.

\bibitem{laird2019soar}
John~E Laird.
\newblock {\em The Soar cognitive architecture}.
\newblock MIT press, 2019.

\bibitem{laird2017standard}
John~E Laird, Christian Lebiere, and Paul~S Rosenbloom.
\newblock A standard model of the mind: Toward a common computational framework across artificial intelligence, cognitive science, neuroscience, and robotics.
\newblock {\em Ai Magazine}, 38(4):13--26, 2017.

\bibitem{lecun2022path}
Yann LeCun.
\newblock A path towards autonomous machine intelligence version 0.9. 2, 2022-06-27.
\newblock {\em Open Review}, 62, 2022.

\bibitem{liang2023taskmatrix}
Yaobo Liang, Chenfei Wu, Ting Song, Wenshan Wu, Yan Xia, Yu~Liu, Yang Ou, Shuai Lu, Lei Ji, Shaoguang Mao, et~al.
\newblock Taskmatrix. ai: Completing tasks by connecting foundation models with millions of apis.
\newblock {\em arXiv preprint arXiv:2303.16434}, 2023.

\bibitem{long2023large}
Jieyi Long.
\newblock Large language model guided tree-of-thought.
\newblock {\em arXiv preprint arXiv:2305.08291}, 2023.

\bibitem{madaan2023self}
Aman Madaan, Niket Tandon, Prakhar Gupta, Skyler Hallinan, Luyu Gao, Sarah Wiegreffe, Uri Alon, Nouha Dziri, Shrimai Prabhumoye, Yiming Yang, et~al.
\newblock Self-refine: Iterative refinement with self-feedback.
\newblock {\em arXiv preprint arXiv:2303.17651}, 2023.

\bibitem{meunier2010modular}
David Meunier, Renaud Lambiotte, and Edward~T Bullmore.
\newblock Modular and hierarchically modular organization of brain networks.
\newblock {\em Frontiers in neuroscience}, 4:200, 2010.

\bibitem{miller2016key}
Alexander Miller, Adam Fisch, Jesse Dodge, Amir-Hossein Karimi, Antoine Bordes, and Jason Weston.
\newblock Key-value memory networks for directly reading documents.
\newblock {\em arXiv preprint arXiv:1606.03126}, 2016.

\bibitem{miller2000prefontral}
Earl~K Miller.
\newblock The prefontral cortex and cognitive control.
\newblock {\em Nature reviews neuroscience}, 1(1):59--65, 2000.

\bibitem{nexus2023}
Nexus.
\newblock Nexusgpt.
\newblock {\em https://nexus.snikpic.io/}, 2023.

\bibitem{openai2023chat}
OpenAI.
\newblock Chatgpt.
\newblock {\em https://chat.openai.com/}, 2022.

\bibitem{chatplugin2023}
OpenAI.
\newblock Chagpt plugins.
\newblock {\em https://openai.com/blog/chatgpt-plugins}, 2023.

\bibitem{openai2023gpt4}
OpenAI.
\newblock {GPT-4} technical report, 2023.

\bibitem{ouyang2022training}
Long Ouyang, Jeffrey Wu, Xu~Jiang, Diogo Almeida, Carroll Wainwright, Pamela Mishkin, Chong Zhang, Sandhini Agarwal, Katarina Slama, Alex Ray, et~al.
\newblock Training language models to follow instructions with human feedback.
\newblock {\em Advances in Neural Information Processing Systems}, 35:27730--27744, 2022.

\bibitem{park2023generative}
Joon~Sung Park, Joseph~C O'Brien, Carrie~J Cai, Meredith~Ringel Morris, Percy Liang, and Michael~S Bernstein.
\newblock Generative agents: Interactive simulacra of human behavior.
\newblock {\em arXiv preprint arXiv:2304.03442}, 2023.

\bibitem{qiao2022reasoning}
Shuofei Qiao, Yixin Ou, Ningyu Zhang, Xiang Chen, Yunzhi Yao, Shumin Deng, Chuanqi Tan, Fei Huang, and Huajun Chen.
\newblock Reasoning with language model prompting: A survey.
\newblock {\em arXiv preprint arXiv:2212.09597}, 2022.

\bibitem{qin2023tool}
Yujia Qin, Shengding Hu, Yankai Lin, Weize Chen, Ning Ding, Ganqu Cui, Zheni Zeng, Yufei Huang, Chaojun Xiao, Chi Han, et~al.
\newblock Tool learning with foundation models.
\newblock {\em arXiv preprint arXiv:2304.08354}, 2023.

\bibitem{reeve2018understanding}
Johnmarshall Reeve.
\newblock {\em Understanding motivation and emotion}.
\newblock John Wiley \& Sons, 2018.

\bibitem{ritter2019act}
Frank~E Ritter, Farnaz Tehranchi, and Jacob~D Oury.
\newblock Act-r: A cognitive architecture for modeling cognition.
\newblock {\em Wiley Interdisciplinary Reviews: Cognitive Science}, 10(3):e1488, 2019.

\bibitem{schick2023toolformer}
Timo Schick, Jane Dwivedi-Yu, Roberto Dess{\`\i}, Roberta Raileanu, Maria Lomeli, Luke Zettlemoyer, Nicola Cancedda, and Thomas Scialom.
\newblock Toolformer: Language models can teach themselves to use tools.
\newblock {\em arXiv preprint arXiv:2302.04761}, 2023.

\bibitem{shao2023synthetic}
Zhihong Shao, Yeyun Gong, Yelong Shen, Minlie Huang, Nan Duan, and Weizhu Chen.
\newblock Synthetic prompting: Generating chain-of-thought demonstrations for large language models.
\newblock {\em arXiv preprint arXiv:2302.00618}, 2023.

\bibitem{shen2023hugginggpt}
Yongliang Shen, Kaitao Song, Xu~Tan, Dongsheng Li, Weiming Lu, and Yueting Zhuang.
\newblock Hugginggpt: Solving ai tasks with chatgpt and its friends in huggingface.
\newblock {\em arXiv preprint arXiv:2303.17580}, 2023.

\bibitem{shinn2023ref}
Noah Shinn, Federico Cassano, Beck Labash, Ashwin Gopinath, Karthik Narasimhan, and Shunyu Yao.
\newblock Reflexion: Language agents with verbal reinforcement learning, 2023.

\bibitem{autogpt2023}
Significant-Gravitas.
\newblock Autogpt.
\newblock {\em https://arxiv.org/pdf/2305.03495.pdf}, 2023.

\bibitem{singer2009researching}
Jerome~L Singer.
\newblock Researching imaginative play and adult consciousness: Implications for daily and literary creativity.
\newblock {\em Psychology of Aesthetics, Creativity, and the Arts}, 3(4):190, 2009.

\bibitem{smallwood2009your}
Jonathan Smallwood, Louise Nind, and Rory~C O’Connor.
\newblock When is your head at? an exploration of the factors associated with the temporal focus of the wandering mind.
\newblock {\em Consciousness and cognition}, 18(1):118--125, 2009.

\bibitem{smallwood2003task}
Jonathan~M Smallwood, Simona~F Baracaia, Michelle Lowe, and Marc Obonsawin.
\newblock Task unrelated thought whilst encoding information.
\newblock {\em Consciousness and cognition}, 12(3):452--484, 2003.

\bibitem{solomon1980opponent}
Richard~L Solomon.
\newblock The opponent-process theory of acquired motivation: the costs of pleasure and the benefits of pain.
\newblock {\em American psychologist}, 35(8):691, 1980.

\bibitem{song2022llm}
Chan~Hee Song, Jiaman Wu, Clayton Washington, Brian~M Sadler, Wei-Lun Chao, and Yu~Su.
\newblock Llm-planner: Few-shot grounded planning for embodied agents with large language models.
\newblock {\em arXiv preprint arXiv:2212.04088}, 2022.

\bibitem{sotelo2020sleep}
Maria~I Sotelo, Jean Tyan, James Dzera, and Ada Eban-Rothschild.
\newblock Sleep and motivated behaviors, from physiology to pathology.
\newblock {\em Current Opinion in Physiology}, 15:159--166, 2020.

\bibitem{stawarczyk2011mind}
David Stawarczyk, Steve Majerus, Michalina Maj, Martial Van~der Linden, and Arnaud D'Argembeau.
\newblock Mind-wandering: Phenomenology and function as assessed with a novel experience sampling method.
\newblock {\em Acta psychologica}, 136(3):370--381, 2011.

\bibitem{vemprala2023chatgpt}
Sai Vemprala, Rogerio Bonatti, Arthur Bucker, and Ashish Kapoor.
\newblock Chatgpt for robotics: Design principles and model abilities.
\newblock {\em Microsoft Autonomous Systems and Robotics Research}, 2023.

\bibitem{wang2023voyager}
Guanzhi Wang, Yuqi Xie, Yunfan Jiang, Ajay Mandlekar, Chaowei Xiao, Yuke Zhu, Linxi Fan, and Anima Anandkumar.
\newblock Voyager: An open-ended embodied agent with large language models.
\newblock {\em arXiv preprint arXiv:2305.16291}, 2023.

\bibitem{wang2023describe}
Zihao Wang, Shaofei Cai, Anji Liu, Xiaojian Ma, and Yitao Liang.
\newblock Describe, explain, plan and select: Interactive planning with large language models enables open-world multi-task agents.
\newblock {\em arXiv preprint arXiv:2302.01560}, 2023.

\bibitem{wei2022chain}
Jason Wei, Xuezhi Wang, Dale Schuurmans, Maarten Bosma, Ed~Chi, Quoc Le, and Denny Zhou.
\newblock Chain of thought prompting elicits reasoning in large language models.
\newblock {\em arXiv preprint arXiv:2201.11903}, 2022.

\bibitem{weiner2013human}
Bernard Weiner.
\newblock {\em Human motivation}.
\newblock Psychology Press, 2013.

\bibitem{wood2012review}
Rachel Wood, Paul Baxter, and Tony Belpaeme.
\newblock A review of long-term memory in natural and synthetic systems.
\newblock {\em Adaptive Behavior}, 20(2):81--103, 2012.

\bibitem{wu2023visual}
Chenfei Wu, Shengming Yin, Weizhen Qi, Xiaodong Wang, Zecheng Tang, and Nan Duan.
\newblock Visual chatgpt: Talking, drawing and editing with visual foundation models.
\newblock {\em arXiv preprint arXiv:2303.04671}, 2023.

\bibitem{xie2023translating}
Yaqi Xie, Chen Yu, Tongyao Zhu, Jinbin Bai, Ze~Gong, and Harold Soh.
\newblock Translating natural language to planning goals with large-language models.
\newblock {\em arXiv preprint arXiv:2302.05128}, 2023.

\bibitem{yao2022react}
Shunyu Yao, Jeffrey Zhao, Dian Yu, Nan Du, Izhak Shafran, Karthik Narasimhan, and Yuan Cao.
\newblock React: Synergizing reasoning and acting in language models.
\newblock {\em arXiv preprint arXiv:2210.03629}, 2022.

\bibitem{baby2023agi}
Yoheinakajima.
\newblock Babyagi.
\newblock {\em https://github.com/yoheinakajima/babyagi}, 2023.

\bibitem{zhang2022automatic}
Zhuosheng Zhang, Aston Zhang, Mu~Li, and Alex Smola.
\newblock Automatic chain of thought prompting in large language models.
\newblock {\em arXiv preprint arXiv:2210.03493}, 2022.

\end{thebibliography}


\clearpage


\end{document}